\documentclass{Interspeech}

\interspeechcameraready

\title{JCAPT: A Joint Modeling Approach for CAPT}

\author[affiliation={}]{Tzu-Hsuan}{Yang}
\author[affiliation={}]{Yue-Yang}{He}
\author[affiliation={}]{Berlin}{Chen}

\affiliation{}{National Taiwan Normal University}{Taipei, Taiwan}

\email{tzuhsuan@ntnu.edu.tw, yueyanghe@ntnu.edu.tw, berlin@ntnu.edu.tw}

\usepackage{comment}
\usepackage{booktabs}
\usepackage{multirow}
\usepackage{array}
\usepackage{graphicx}
\usepackage{placeins}

\begin{document}

\maketitle

\begin{abstract}
Effective pronunciation feedback is critical in second language (L2) learning, for which computer-assisted pronunciation training (CAPT) systems often encompass two key tasks: automatic pronunciation assessment (APA) and mispronunciation detection and diagnosis (MDD). Recent work has shown that joint modeling of these two tasks can yield mutual benefits. Our unified framework leverages Mamba, a selective state space model (SSM), while integrating phonological features and think token strategies to jointly enhance interpretability and fine-grained temporal reasoning in APA and MDD. To our knowledge, this is the first study to combine phonological attribution, SSM-based modeling, and prompting in CAPT. A series of experiments conducted on the speechocean762 benchmark demonstrate that our model consistently outperforms prior methods, particularly on the MDD task.
\end{abstract}

\section{Introduction}

In the era of globalized communication, learning a second language (L2) has become increasingly essential. Computer-assisted pronunciation training (CAPT) systems have emerged as practical and scalable solutions. These systems provide learners with a low-pressure, self-directed environment to enhance their pronunciation skills through immediate, objective, and personalized feedback \cite{Munday17-Duo, Kholis21-Elsa}. To offer meaningful guidance, effective CAPT systems are typically composed of two integral components: automatic pronunciation assessment (APA), which delivers a broader evaluation of speaking skills; and mispronunciation detection and diagnosis (MDD), which aims to identify pronunciation errors and provide detailed diagnostic feedback.

A de-facto CAPT system typically operates in a read-aloud scenario, where L2 learners are prompted to speak predefined sentences. In this context, APA modules assess pronunciation quality across various aspects (e.g., accuracy, fluency, and completeness) and multiple linguistic granularities (e.g., phoneme-, word-, and utterance-levels) \cite{Gong22-MultiAspect, Yan23-OrdinalLoss, yan2024effective}. On a separate front, MDD focuses on identifying phonetic pronunciation errors commonly made by non-native speakers \cite{wang2022exploring, Ye22-APL, Yan23-Peppanet}. Such errors tend to have clear-cut distinctions between correct and incorrect pronunciations and can be systematically identified via phoneme-level mismatches including deletions, substitutions, and insertions. These two components collaboratively enable CAPT systems to deliver holistic and pedagogically relevant feedback to L2 learners for improving their pronunciation proficiency.

Recent research has shown that integrating APA and MDD into a unified modeling framework can enhance the performance of both tasks \cite{ryu2023joint, he2024jam, chao2025towards}. Such a synergy enhances not only the granularity of mispronunciation detection but also the reliability of multi-aspect pronunciation scoring. 
However, an underexplored, yet promising direction involves the use of articulatory attributes, which provide a linguistically grounded representation of phoneme realization, such as voicing, manner, place of articulation, and others. While these articulatory attributes have been leveraged to improve MDD performance \cite{Shahin24-PhonologicalMD, yan2023effective}, their role in enhancing multi-aspect APA, in conjunction with MDD and under a joint modeling framework, remains largely unexplored.

Beyond incorporating linguistic knowledge, architectural innovations can also significantly affect the effectiveness of CAPT systems. In particular, increasing the capacity of model components to perform deeper reasoning or richer representation learning at the frame- or linguistic token-levels has shown promise. Inspired by chain-of-thought reasoning in natural language processing \cite{wei2022chain}, the concept of “think tokens” has been proposed to increase the computational depth per token. This idea has been applied to Transformer-based ASR models \cite{yang2024contemplative}, where encouraging frame-wise reflection has improved both inference quality and training efficiency. In parallel, Mamba \cite{gu2023mamba}, a selective state space model (SSM), has demonstrated strong potential for modeling long-range dependencies with high computational efficiency. These characteristics make it a promising candidate for fine-grained speech assessment tasks \cite{chao2025towards}. There are some research efforts exploring how prompting mechanisms can be adapted to SSM-based architectures such as Mamba \cite{yoshimura2024mambapeft}, but to our knowledge, there is little work focusing on its applicability in CAPT scenarios, especially for phoneme-level representations.

\setlength{\abovecaptionskip}{-3pt}
\begin{figure*}[htbp]
\begin{center}
\includegraphics[width=\linewidth, keepaspectratio]{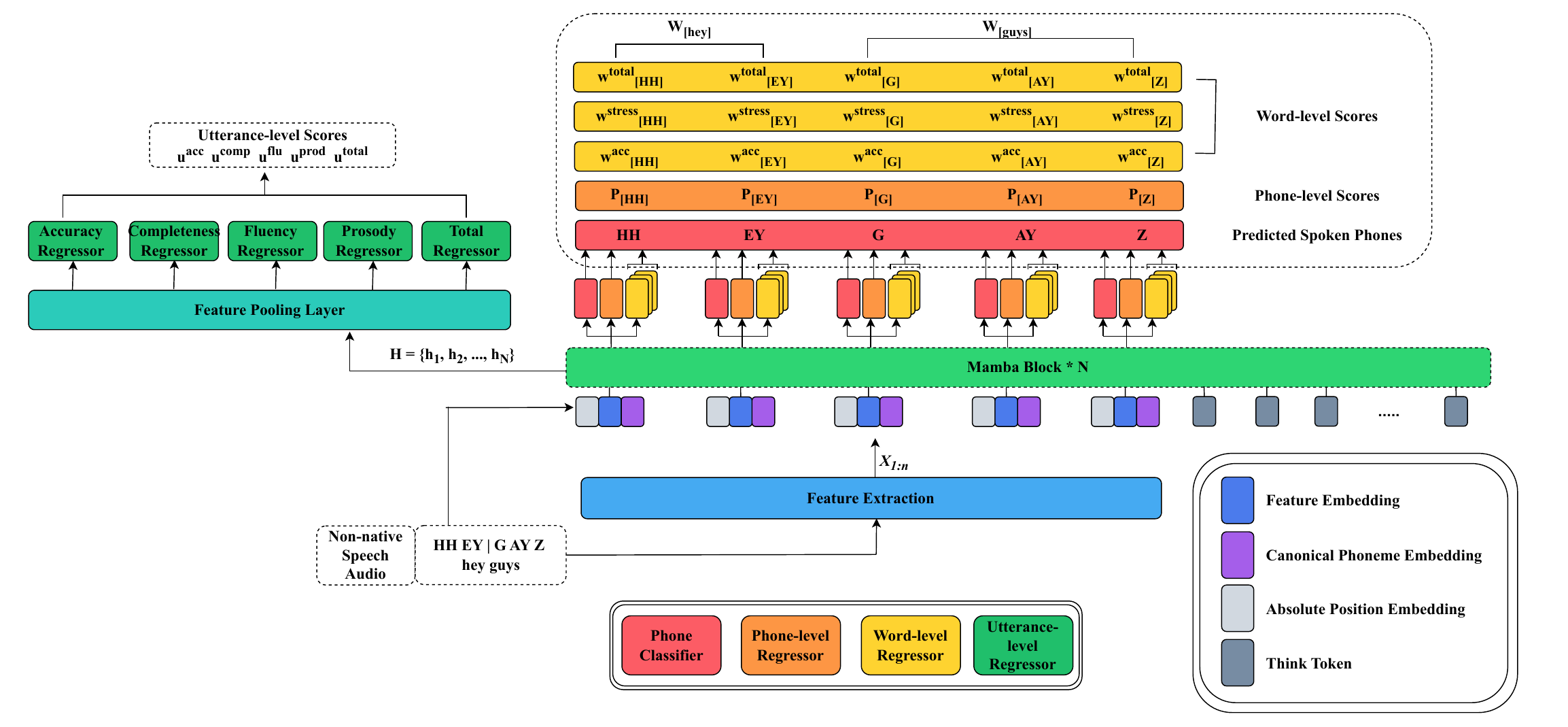}
\end{center}
\caption{The overall architecture of the multi-task learning model for pronunciation assessment and mispronunciation detection and diagnosis. The system consists of feature extraction, Mamba layers for temporal modeling, and multi-level scoring modules including phoneme-level, word-level, and utterance-level regressors. }
\label{main_architecture}
\end{figure*}

In this paper, we present JCAPT, a \textbf{J}oint \textbf{CAPT} framework that jointly considers APA and MDD tasks using a parallel architecture, as illustrated in Figure~\ref{main_architecture}. JCAPT leverages Mamba, a state space model (SSM) capable of capturing long-range temporal dependencies with high computational efficiency, for rendering phone- and word-level pronunciation characteristics. In addition to utilizing canonical phone information as in prior work \cite{Gong22-MultiAspect, yan2023preserving, he2024jam}, our model also integrates phonological features to enhance diagnostic precision and interpretability. To our knowledge, our work is among the first to investigate the synergistic effect of phonological attribution, think token strategies, and Mamba-based architectures within a unified modeling framework for assessing the pronunciation proficiency of L2 learners. We validate the proposed JCAPT approach through comprehensive experiments on the speechocean762 benchmark, demonstrating that JCAPT significantly outperforms existing baselines, especially with respect to MDD performance. Notably, we find that completeness—an aspect of APA that remains particularly challenging—can be substantially improved by subtler modeling of phoneme-level pronunciation traits.

\section{Methodology}
Figure~\ref{main_architecture} schematically visualizes our proposed framework, JCAPT, which jointly models Automatic Pronunciation Assessment (APA) and Mispronunciation Detection and Diagnosis (MDD) through a parallel architecture. Our system consists of five key components: 
1) a comprehensive feature extraction module that integrates multiple speech representations; 
2) a bi-directional Mamba encoder for contextual modeling; 
3) a contemplative reasoning mechanism via think tokens; 
4) an attention-based pooling layer; 
and 5) multi-level scoring heads for APA and MDD. 

\begin{table*}[t]
\centering
\setlength{\tabcolsep}{2.5pt} 
\renewcommand{\arraystretch}{1} 
\resizebox{\linewidth}{!}{%
\begin{tabular}{lccccc ccccc ccc}
\toprule
\multirow{2}{*}{\textbf{Models}} & \multicolumn{2}{c}{\textbf{Phoneme-level}} & \multicolumn{3}{c}{\textbf{Word-level}} & \multicolumn{5}{c}{\textbf{Utterance-level}} & \multicolumn{3}{c}{\textbf{MDD Performance}} \\
\cmidrule(lr){2-3} \cmidrule(lr){4-6} \cmidrule(lr){7-11} \cmidrule(lr){12-14}
& MSE ↓ & PCC ↑ & Acc. ↑ & Stress ↑ & Total ↑ & Acc. ↑ & Comp. ↑ & Fluency ↑ & Prosody ↑ & Total ↑ & RE. (\%) ↑ & PR. (\%) ↑ & F1. (\%) ↑ \\
\midrule
Joint-CAPT-L1 \cite{ryu2023joint} & - & - & - & - & - & 0.719 & - & 0.775 & 0.773 & 0.743 & \textbf{91.40} & 26.70 & 41.40 \\
JAM \cite{he2024jam}                     & 0.076 & 0.664 & 0.622 & 0.241 & 0.638 & 0.773 & 0.205 & 0.831 & \textbf{0.829} & 0.805 & 34.76 & 64.10 & 45.01 \\
JCAPT                   & \textbf{0.066} & \textbf{0.720} & \textbf{0.699} & \textbf{0.270} & \textbf{0.711} & \textbf{0.783} & \textbf{0.551} & \textbf{0.834} & 0.824 & \textbf{0.806} & 40.23 & \textbf{69.89} & \textbf{51.05} \\
\bottomrule
\end{tabular}
}
\caption{Experimental results of different methods evaluated on speechocean762. \text{Acc. and Comp.} refers to Accuracy and Completeness, respectively. }
\label{tab:main_results}
\end{table*}
\subsection{Feature Extraction}
Given a speech utterance from an L2 learner and the canonical phone sequence $p = \{p_1, p_2, \ldots, p_N\}$ of the corresponding text prompt, our system extracts a set of phone-level features by combining goodness of pronunciation (GOP) \cite{witt2000phone, hu2015improved, shi2020context} with modern self-supervised representations. 

\textbf{Goodness of Pronunciation:}
GOP is a widely used feature that measures the likelihood of each phone being correctly pronounced by a speaker. We follow the standard pipeline for GOP computation, which includes forced alignment via a DNN-HMM acoustic model, canonical phoneme decoding, and posterior probability estimation. This produces phone-aligned GOP features that directly reflect pronunciation accuracy. Due to its explicit modeling of phoneme correctness, GOP is particularly effective for MDD.

\textbf{Self-Supervised Representations:}
To capture rich contextual and articulatory information, we incorporate three self-supervised learning (SSL) models: wav2vec 2.0 \cite{baevski2020wav2vec}, HuBERT \cite{hsu2021hubert}, and WavLM \cite{chen2022wavlm}, which are pre-trained on large-scale unlabeled speech data. We extract frame-level hidden features from each SSL model and align them to the canonical phone boundaries using forced alignment. The aligned features are then concatenated with GOP scores to form a comprehensive phoneme-level feature vector. The resulting feature vectors of an input utterance are projected through a dense layer to obtain a sequence of phone-level embeddings, denoted as $x_{1:N}$, where $N$ is the number of canonical phones.

\textbf{Canonical Phone Embedding:}
In parallel, we construct canonical phone embeddings to provide symbolic linguistic supervision. For each phoneme $p_i$, we convert it into a one-hot vector $Phn_{\text{onehot}}$, and concatenate it with a phonological attribute vector $Phn_{\text{attr}}$ that encodes articulatory properties. The resulting phoneme-level symbolic representation is projected into the same dimension as $x_{1:N}$, and later fused into the model as an auxiliary input.

\subsection{Bi-directional Mamba Encoder}
To model long-range dependencies with low complexity, we adopt a bi-directional Mamba encoder inspired by the Dual-Mamba architecture \cite{jiang2025dual}. Its linear scaling and efficient temporal representation make it well-suited for fine-grained speech assessment. 

Given the phoneme-level acoustic feature sequence $x_{1:N}$ and the canonical phoneme embedding sequence that is constructed from the concatenation of phoneme one-hot vectors and their associated phonological attributes, we fuse both inputs to form the final encoder input:
\begin{equation}
    \hat{x}_i = x_i + c_i, \quad \text{for } i = 1, \ldots, N
\end{equation}
where $x_i$ is the projected acoustic feature and $c_i$ is the symbolic canonical embedding.

The resulting sequence $\hat{X} = \{\hat{x}_1, \hat{x}_2, \ldots, \hat{x}_N\}$ is then passed through a stack of bidirectional Mamba blocks. To enhance the encoder’s reasoning capacity, we append a set of learnable \textit{think tokens} to the input sequence. Their mechanism and motivation are detailed in Section~\ref{sec:contemplative_reasoning}. 

The output of the encoder is a sequence of contextualized phoneme-level representations:
\begin{equation}
    H = \mathrm{BiMamba}(\hat{X}, Emb_{think}) = \{h_1, h_2, \ldots, h_N\}, 
\end{equation}
where $Emb_{think}$ denotes the embedding of the think tokens.
This enriched representation $H$ forms the basis for the subsequent reasoning and prediction modules.

\subsection{Contemplative Reasoning via Think Tokens}
\label{sec:contemplative_reasoning}
Inspired by contemplative prompting \cite{yang2024contemplative}, we introduce \textit{think tokens} to encourage deeper reasoning over each phoneme-level representation. 
Instead of inserting interleaved think tokens within a Transformer-based architecture, we postpend a fixed number of think tokens to the end of the input sequence, allowing the model to perform additional internal computation before making phoneme-level predictions. These tokens are implemented as learnable embeddings and are jointly optimized during training. This mechanism is particularly effective in enhancing the diagnostic capacity of MDD and improving the consistency of multi-aspect APA predictions.

\subsection{Attention-based Feature Pooling}
To obtain utterance-level representations for multi-aspect pronunciation assessment, we employ a set of aspect-specific attention-based pooling mechanisms. Given the encoder output $H = \{h_1, h_2, \ldots, h_N\} \in \mathbb{R}^{N \times d}$, we define a separate attention module for each assessment aspect $a \in \mathcal{A}$, where $\mathcal{A}$ denotes the set of predefined aspects (e.g., accuracy, fluency, prosody).

For each aspect $a$, attention weights are computed by
\begin{equation}
    \alpha_i^{(a)} = \frac{\exp\left(w_a^\top \tanh(W_a h_i)\right)}{\sum_{j=1}^{N} \exp\left(w_a^\top \tanh(W_a h_j)\right)},
\end{equation}
where $W_a \in \mathbb{R}^{d_a \times d}$ and $w_a \in \mathbb{R}^{d_a}$ are learnable parameters for aspect $a$.

The utterance-level representation for aspect $a$ is obtained by:
\begin{equation}
    h_u^{(a)} = \sum_{i=1}^{N} \alpha_i^{(a)} h_i.
\end{equation}

This design allows each aspect to focus on different parts of the input sequence, reflecting its unique contribution to overall pronunciation quality.

\subsection{Multi-level Scoring Heads}
To support both the APA and MDD tasks across different granularities, we design hierarchical prediction heads. For each contextualized phoneme representation $h_i$, a regression head estimates phoneme-level APA scores, while a classification head predicts corresponding MDD labels. Word-level scores are derived by aggregating phoneme-level outputs based on forced alignment boundaries. For utterance-level APA, each $h_u^{(a)}$ obtained from the attention-based pooling layer is passed through an individual regression head corresponding to the aspect $a$, yielding holistic multi-aspect assessment scores. This multi-level design enables joint modeling of fine-grained and global pronunciation quality indicators.

\subsection{Optimization}
Our model is trained under a multi-task learning (MTL) framework that optimizes both the APA and MDD objectives jointly. 
For APA, we formulate the objective as the sum of losses across different granularity levels:
\begin{equation}
    \mathcal{L}_{\text{APA}} = \mathcal{L}_{\text{phn}} + \mathcal{L}_{\text{word}} + \mathcal{L}_{\text{utt}}
    \label{eq:loss_capt},
\end{equation}
where each component denotes the mean squared error (MSE) loss at the phoneme-, word-, and utterance-levels, respectively. 

For MDD, a phoneme-level classifier is trained using cross-entropy loss to improve mispronunciation detection:
\begin{equation}
\mathcal{L}_{\text{MDD}} = -\sum_{i=1}^{N} \sum_{p=1}^{P} y_{i,p} \log(\hat{y}_{i,p})
\label{eq:loss_mdd},
\end{equation}
where $N$ is the number of training instances, $P$ is the number of phoneme classes, $y_{i,p}$ is the one-hot ground truth, and $\hat{y}_{i,p}$ is the predicted probability for the $p$-th phoneme in the $i$-th instance.

The ultimate loss is a weighted combination of the CAPT and MDD losses:
\begin{equation}
\mathcal{L} = (1 - \alpha) \cdot \mathcal{L}_{\text{APA}} + \alpha \cdot \mathcal{L}_{\text{MDD}}
\label{eq:loss_total},
\end{equation}
where $\alpha$ strikes a balance between these two objectives. In our implementation, we set $\alpha = 0.3$ following \cite{he2024jam}, which was found to yield stable performance in similar settings.

\begin{table*}[t]
\centering
\small
\setlength{\tabcolsep}{5pt}
\renewcommand{\arraystretch}{1.2}
\resizebox{\linewidth}{!}{%
\begin{tabular}{l cc ccc ccccc}
\toprule
\multirow{2}{*}{\textbf{Models}} & \multicolumn{2}{c}{\textbf{Phoneme-level Score}} & \multicolumn{3}{c}{\textbf{Word-level Score (PCC)}} & \multicolumn{5}{c}{\textbf{Utterance-level Score (PCC)}} \\
\cmidrule(lr){2-3} \cmidrule(lr){4-6} \cmidrule(lr){7-11}
 & MSE ↓ & PCC ↑ & Acc. ↑ & Stress ↑ & Total ↑ & Acc. ↑ & Comp. ↑ & Fluency ↑ & Prosody ↑ & Total ↑ \\
\midrule
\textbf{JCAPT}                             & \textbf{0.066} & \textbf{0.720} & \textbf{0.699} & 0.270          & \textbf{0.711} & 0.783          & 0.551          & 0.834          & 0.824          & 0.806          \\
w/o phonological                     & \textbf{0.066} & 0.716          & 0.689          & 0.239          & 0.701          & 0.775          & \textbf{0.644} & \textbf{0.840} & \textbf{0.826} & \textbf{0.808} \\
w/o think tokens                        & \textbf{0.066} & \textbf{0.720} & \textbf{0.699} & \textbf{0.309} & 0.710          & \textbf{0.784} & 0.556          & 0.833          & 0.818          & \textbf{0.808} \\
w/o phonological, think token     & 0.068          & 0.708          & 0.687          & 0.273          & 0.699          & 0.779          & 0.547          & 0.834          & 0.822          & \textbf{0.808} \\
\bottomrule
\end{tabular}
}
\caption{Ablation Studies of the proposed method on automatic pronunciation assessment.}
\label{abalation_APA}
\end{table*}

\begin{table*}[t]
\centering
\small
\setlength{\tabcolsep}{8pt} 
\renewcommand{\arraystretch}{1.2}
\begin{tabular}{lccccc}
\toprule
\multicolumn{1}{c}{\multirow{2}{*}{\textbf{Models}}} & \multicolumn{4}{c}{\textbf{MDD results}} & \textbf{Diagnosis results} \\
\cmidrule(lr){2-5} \cmidrule(lr){6-6}
\multicolumn{1}{c}{} & Recall (\%) ↑ & Precision (\%) ↑ & F1-score (\%) ↑ & PER (\%) ↓ & Correct Diag. (\%) ↑ \\
\midrule
JCAPT                                 & 40.23 & 69.89 & 51.05 & \textbf{2.66} & \textbf{54.42} \\
w/o phonological                 & \textbf{42.00} & 68.98 & \textbf{52.21} & 2.70 & 52.67 \\
w/o think tokens                    & 39.76 & 69.95 & 50.61 & 2.67 & 54.35 \\
w/o phonological, think token & 41.27 & \textbf{70.07} & 51.92 & 2.67 & 52.80 \\
\bottomrule
\end{tabular}
\caption{Abalation Studies of proposed method on mispronunciation detection and diagnosis.}
\label{ablation_MDD}
\end{table*}

\section{Experiments and Results}
\subsection{Dataset}
We conducted our experiments on the speechocean762 dataset \cite{zhang2021speechocean762}, a publicly available benchmark designed for research on automatic pronunciation assessment (APA) and mispronunciation detection and diagnosis (MDD). The dataset contains 5,000 English utterances produced by 250 Mandarin-speaking L2 learners, evenly divided into training and test sets. 
Each utterance is annotated with human-rated pronunciation scores at the utterance, word, and phoneme levels, assessed by five expert raters using standardized rubrics. For the MDD task, the dataset provides canonical and realized phone-level transcriptions, aligned at the phoneme level. It adopts a 39-phone set, based on the CMU pronunciation dictionary \cite{weide2005carnegie} and extended with $<$del$>$ and $<$unk$>$ tokens to indicate deleted and noncategorizable phones. In particular, the data set does not include insertion errors, which facilitates cleaner alignment between canonical and observed pronunciations. 
\subsection{Experimental Setup}
Following the experimental configuration outlined in \cite{he2024jam}, we adopted the same procedures for extracting both GOP and SSL features. Additionally, we incorporated phonological attributes as introduced in \cite{Shahin24-PhonologicalMD}, enriching our phoneme-level representations. To ensure robustness and reproducibility, we conducted five independent runs with different random seeds. In the following experiments, we will report the average performance in terms of Pearson Correlation Coefficient (PCC) and Mean Squared Error (MSE) across all runs. 

In addition, following previous MDD studies \cite{li2017mispronunciation}, we adopt F1-score, the harmonic mean of recall (RE.) and precision (PR.), as the primary metric for evaluating mispronunciation detection performance. Furthermore, we report the phoneme error rate (PER) to reflect overall detection accuracy at the segmental level. For diagnostic evaluation, we include the correct diagnosis rate (Correct Diag.), which measures the proportion of detected errors that are correctly classified, providing insight into the interpretability and reliability of the system's feedback.

\subsection{Main Results}
As shown in Table~\ref{tab:main_results}, JCAPT consistently outperforms previous models across all evaluation levels on the speechocean762 benchmark. At the phoneme level, it achieves the lowest MSE and highest PCC, indicating more accurate phoneme-level scoring achieved. Furthermore, most word-level assessment performance is boosted, with notable gains in stress prediction and total accuracy over JAM~\cite{he2024jam}. For the utterance-level assessment, JCAPT reports higher completeness, fluency, and prosody scores, reflecting enhanced modeling of global speech characteristics. In MDD, our model significantly boosts recall and F1-score, demonstrating superior detection of mispronunciations. These results confirm that integrating phonological features, think token mechanisms, and the Mamba architecture yields more accurate and interpretable CAPT performance.

\subsection{Ablation Studies}
To evaluate the contribution of each module, we conducted ablation studies under three settings: (1) removing phonological features, (2) removing the think tokens, and (3) removing both. As shown in Tables~\ref{abalation_APA} and~\ref{ablation_MDD}, the full model consistently outperforms all ablated versions in both the APA and MDD tasks.

First, removing phonological features leads to notable drops in phoneme- and word-level performance, especially in MSE, PCC, and stress prediction, highlighting their importance for modeling fine-grained articulatory patterns. In contrast, utterance-level metrics remain stable, suggesting limited influence on global prosodic traits. Second, excluding the think tokens mainly affects MDD, reducing recall and F1, which implies its effectiveness in capturing disfluency-related cues. However, minor decreases in precision indicate a potential trade-off in prediction stability. 
Lately, when both components are removed, performance degrades across the board, suggesting that they may play complementary roles: phonological features might provide linguistic grounding, while the think tokens may enhance cognitive sensitivity in pronunciation modeling.

\section{Conclusion and Future Work}
In this work, we have put forward JCAPT, a unified CAPT framework that jointly addresses APA and MDD through a parallel architecture built upon the Mamba state space model. By integrating phonological features and adopting a “think token” strategy for fine-grained temporal reasoning, JCAPT enhances both diagnostic interpretability and predictive performance. Our experimental results on the speechocean762 benchmark show that JCAPT outperforms baselines, especially in mispronunciation detection and completeness—highlighting the value of joint modeling and phoneme-level reasoning in L2 pronunciation assessment. As to future work, we will explore the generalizability of our framework across diverse learner populations, languages, and spontaneous speech scenarios. In addition, we intend to implement item-specific enhancements for scoring aspects with relatively lower correlation coefficients, particularly stress and completeness, to address their current limitations and improve the overall robustness and modeling granularity of our CAPT system.

\section{Acknowledgement}
This work was supported by the Language Training and Testing Center (LTTC), Taiwan. Any findings and implications in the paper do not necessarily reflect those of the sponsor.

\bibliographystyle{IEEEtran}
\bibliography{mybib}

\end{document}